\documentclass[runningheads]{llncs}

\usepackage{eccv}

\usepackage{eccvabbrv}
\usepackage{xcolor}
\usepackage{graphicx}
\usepackage{booktabs}
\usepackage{tabu}
\usepackage{amsmath}
\usepackage{amssymb}
\usepackage{bm}
\usepackage{verbatim}
\usepackage{multirow}
\usepackage{gensymb}
\usepackage{color}
\usepackage{marvosym}
\usepackage{colortbl}
\usepackage{makecell}
\usepackage[normalem]{ulem}
\usepackage{balance}
\usepackage{arydshln}
\usepackage{setspace}
\usepackage[accsupp]{axessibility}  
\usepackage[marginal]{footmisc}
\usepackage[ruled]{algorithm2e}
\usepackage[accsupp]{axessibility}  

\usepackage{fontawesome}
\usepackage{hyperref}

\usepackage{orcidlink}

\begin{document}

\title{PolyRoom: Room-aware Transformer for Floorplan Reconstruction} 


\author{Yuzhou Liu\inst{1,2,3}\orcidlink{0009-0002-3334-2982} \and
Lingjie Zhu\inst{4} \and
Xiaodong Ma\inst{5} \and
Hanqiao Ye\inst{1,2,3}\orcidlink{0009-0004-0313-7085} \and
Xiang Gao\inst{1,3}\textsuperscript{\faEnvelopeO} \orcidlink{0000-0003-1497-5637} \and
Xianwei Zheng\inst{6}\orcidlink{0000-0001-9783-3030} \and
Shuhan Shen\inst{1,2,3}\textsuperscript{\faEnvelopeO} \orcidlink{0000-0002-8704-7914}}

\footnotetext{\textsuperscript{\faEnvelopeO}Corresponding authors}
\authorrunning{Y. Liu et al.}

\institute{Institute of Automation, Chinese Academy of Sciences \and
School of Artificial Intelligence, University of Chinese Academy of Sciences \and
\mbox{CASIA-SenseTime Research Group \and Cenozoic Robotics} \and \mbox{Ocean University of China \and
The State Key Lab. LIESMARS, Wuhan University} \\ 
\email{\{liuyuzhou2021,yehanqiao2022,xiang.gao\}@ia.ac.cn; lingjie.zhu.me@gmail.com;  maxiaodong5000@stu.ouc.edu.cn; zhengxw27@foxmail.com; shshen@nlpr.ia.ac.cn}}
\maketitle
\begin{figure}[!h] 
\centering
\includegraphics[width=0.9\linewidth]{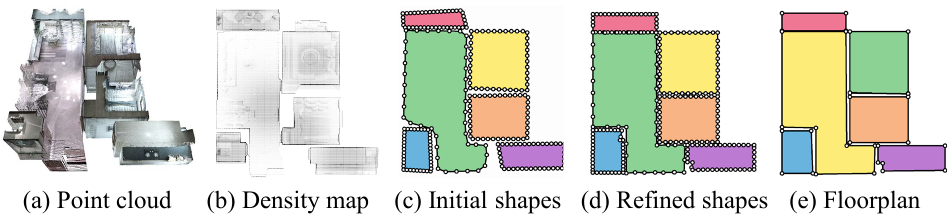}
\caption{\textbf{Structured floorplan reconstruction by  PolyRoom.} Given an indoor point cloud (a), PolyRoom predicts the initial room shapes (c) from the density map (b). Then, PolyRoom refines them gradually (d) and reconstructs the final vectorized floorplan (e).}
\label{pic2}
\end{figure}
\begin{abstract}
Reconstructing geometry and topology structures from raw unstructured data has always been an important research topic in indoor mapping research. 
In this paper, we aim to reconstruct the floorplan with a vectorized representation from point clouds. 
Despite significant advancements achieved in recent years, current methods still encounter several challenges, such as missing corners or edges, inaccuracies in corner positions or angles, self-intersecting or overlapping polygons, and potentially implausible topology. 
To tackle these challenges, we present PolyRoom, a room-aware Transformer that leverages uniform sampling representation, room-aware query initialization, and room-aware self-attention for floorplan reconstruction. 
Specifically, we adopt a uniform sampling floorplan representation to enable dense supervision during training and effective utilization of angle information. Additionally, we propose a room-aware query initialization scheme to prevent non-polygonal sequences and introduce room-aware self-attention to enhance memory efficiency and model performance. 
Experimental results on two widely used datasets demonstrate that PolyRoom surpasses current state-of-the-art methods both quantitatively and qualitatively. 
Our code is available at: \url{https://github.com/3dv-casia/PolyRoom/}.
\end{abstract}
\section{Introduction}

Structured modeling~\cite{ikehata2015structured, zhang2021structured} aims to represent a scene using basic geometric primitives while maintaining the complete topology and precise geometry holistically. Different from other commonly employed representations such as point clouds, meshes, or voxels, structured representations are lightweight and editable, facilitating human scene understanding and downstream applications, such as VR/AR, robotics, and navigation~\cite{li2021cognitive,liu2024lightweight,chen2024f3loc}. However, efficiently reconstructing structures from raw unstructured data remains an open problem.
\par
For structured indoor modeling, various representations are utilized, ranging from 2D layouts~\cite{lee2017roomnet} and 3D wireframes~\cite{sun2019horizonnet} to 2D floorplans~\cite{chen2019floor}, aiming to reconstruct scenes from diverse data sources such as single monocular RGB images~\cite{ibrahem2023st}, panoramic images~\cite{jiang2022lgt}, CAD drawings~\cite{zheng2022gat}, and point clouds~\cite{cabral2014piecewise}.
Among different input data and output representations, vectorized floorplan exhibit significant advantages in accurately depicting scene geometric details and global topology structure, while point clouds are deemed to be more geometrically precise and contain rich visual information. 
Consequently, the trend of using point clouds as input and predicting vectorized 2D floorplan has gained popularity in indoor mapping~\cite{chen2019floor,yue2023connecting}.
However, in the face of noise, incompleteness, and other inherent challenges in raw point clouds, floorplan reconstruction remains a demanding task. 

To address the challenges of reconstructing structured floorplan from 3D point clouds, there has been a steady paradigm shift from traditional geometric optimization methods to end-to-end inference approaches. 
Traditional optimization-based methods involve fitting planes from point clouds and obtaining the final floorplan through optimization techniques~\cite{cabral2014piecewise, han2023floorusg}. 
To simplify the problem, it is a common practice to project the 3D indoor point clouds along the gravity axis and proceed with the resulting 2D density map.  
Some approaches propose combining deep learning and geometric optimization techniques by initially detecting room proposals and then solving it as an optimization problem~\cite{chen2019floor, stekovic2021montefloor}, which are not end-to-end and rely on prior-based optimization method design. Recently, with Transformer~\cite{vaswani2017attention} showcasing remarkable capabilities in sequence modeling~\cite{liu2021swin}, researchers have attempted to adopt it for indoor mapping tasks, and it also yields promising results~\cite{chen2022heat,yue2023connecting}.

For the Transformer-based methods,  the floorplan representation has critical impacts on the choice of techniques and the organization of the algorithm.
HEAT~\cite{chen2022heat} represents floorplan as discrete corners and edges. However, this representation suffers from missing corners and edges due to missed detection. SLIBO-Net~\cite{su2023slibo} adopts a slicing box representation but is limited by the Manhattan assumption. Representing floorplan as polygon sequences eliminates reliance on the Manhattan assumption. RoomFormer~\cite{yue2023connecting} first introduces the two-level queries, which depict the floorplan as variable-size polygons represented by variable-length ordered corners. Despite significant improvements in reconstruction results, RoomFormer~\cite{yue2023connecting} still faces challenges such as disordered sequences from random query initialization, as well as inaccuracies in polygons due to missing or biased single corners. Overall, current methods still encounter challenges with incomplete and inaccurate reconstruction results. \par 

In this paper, we propose a novel floorplan reconstruction method named PolyRoom, designed to address floorplan reconstruction as a refinement process of room queries from initial shapes (as depicted in Figure~\ref{pic2}). 
PolyRoom introduces the uniform sampling floorplan representation, enabling room-aware query initialization with fixed-length vertex sequences.  Furthermore, it incorporates dense supervision during training and utilizes angles to enhance performance.
To mitigate inaccurate polygon sequences resulting from randomly initialized learnable queries, PolyRoom introduces room-aware query initialization to align the queries with room semantics.
With the initialization, the initial shapes of room polygons are refined layer by layer in the Transformer decoder. Considering the memory overhead of self-attention, PolyRoom utilizes room-aware self-attention to achieve better performance with reduced memory consumption.

We evaluate PolyRoom on two widely adopted datasets, Structured3D~\cite{zheng2020structured3d} and SceneCAD~\cite{avetisyan2020scenecad}. On both datasets, PolyRoom surpasses the state of the arts. Besides, PolyRoom also shows better generalization ability. We outline our contributions below:\par
\begin{itemize}
\item We propose PolyRoom, a novel method that views floorplan reconstruction as a progressive refinement of initial room shapes in an encoder-decoder architecture.
\item The uniform sampling representation depicts room polygons as fixed-length vertex sequences, providing various dense supervision during training and facilitating the introduction of room-aware query initialization.
\item Room-aware query initialization utilizing instance segmentation yields superior initial queries compared with randomly initialized learnable queries, enhancing the accuracy of predicted polygon sequences.
\item Room-aware self-attention incorporates intra-room and inter-room self-attention based on the floorplan representation, reducing memory consumption while achieving better reconstruction results.
\end{itemize}
\section{Related Work}
\subsection{Vectorized Floorplan Reconstruction} Floorplan reconstruction entails generating a structured representation of indoor scenes from unstructured data. Various approaches have been explored in this field. Some researchers fit planes from point clouds and select lines to form the floorplan using optimization techniques~\cite{cabral2014piecewise,han2023floorusg}. Others infer floorplan from panoramic images~\cite{sun2019horizonnet,jiang2022lgt} or CAD drawings~\cite{fan2021floorplancad,zheng2022gat} in end-to-end manners. In recent years, there has been a growing interest in floorplan reconstruction with density maps generated by point cloud projection. Some methods integrate instance segmentation with optimization algorithms to address this challenge~\cite{chen2019floor,stekovic2021montefloor}. However, they are constrained by the prior-based optimization function design and long optimization time. Given the remarkable success of applying  Transformer~\cite{vaswani2017attention} in computer vision~\cite{liu2021swin,dosovitskiy2020image}, floorplan reconstruction has also embraced its adoption.
HEAT~\cite{chen2022heat} uses a Transformer to detect corners and infer connection relationships between them to create floorplan planar graph, but it may fail when certain corners are absent, leading to missing edges and unclosed rooms. RoomFormer~\cite{yue2023connecting} represents floorplan as two-level queries, predicting the polygon sequences to ensure the room closeness. Nonetheless, it faces challenges such as incorrect polygon sequences and distorted polygon contours. Recently, PolyDiffuse~\cite{chen2023polydiffuse} first uses diffusion models to generate refined floorplan based on initial proposals or human annotations. SLIBO-Net\cite{su2023slibo} employs the slicing box representation and introduces a regularization mechanism and post-processing step with geometric priors to capture finer local geometric details.

\subsection{Polygonal Instance Segmentation}
Floorplan reconstruction using polygon sequence representation treats rooms as polygons, akin to polygonal instance segmentation which predicts polygonal instance contours. Some researchers extract polygon instances based on geometric primitives. For instance, Delaunay point processes~\cite{favreau2019extracting} embed point processes into a Delaunay triangulation and utilize its key properties in a Markov Chain Monte Carlo sampler~\cite{browne2012survey} to extract geometric structures from images. 
Besides the optimization-based methods, an increasing number of end-to-end polygonal instance segmentation algorithms have emerged, which typically employ time-series models to predict polygon boundaries. Castrejon~\etal~\cite{castrejon2017annotating} uses LSTM~\cite{hochreiter1997long} to sequentially produce polygon vertices of object contours.  PolyTransform~\cite{liang2020polytransform} employs a segmentation network to generate instance masks and refines initial contours with a deforming network to better fit the object contours. DeepSnake~\cite{peng2020deep} implements the snake algorithm by iteratively deforming an initial contour to match the object boundary. BoundaryFormer~\cite{lazarow2022instance} utilizes a Transformer architecture to directly predict polygon contours using a mask-based loss through a differentiable rasterizer. Recently, PolyFormer~\cite{liu2023polyformer} predicts the sequence of polygon vertices with large language models by combining image patches and text query tokens.
\begin{figure*}[!t] 
\centering
\includegraphics[width=0.95\linewidth]{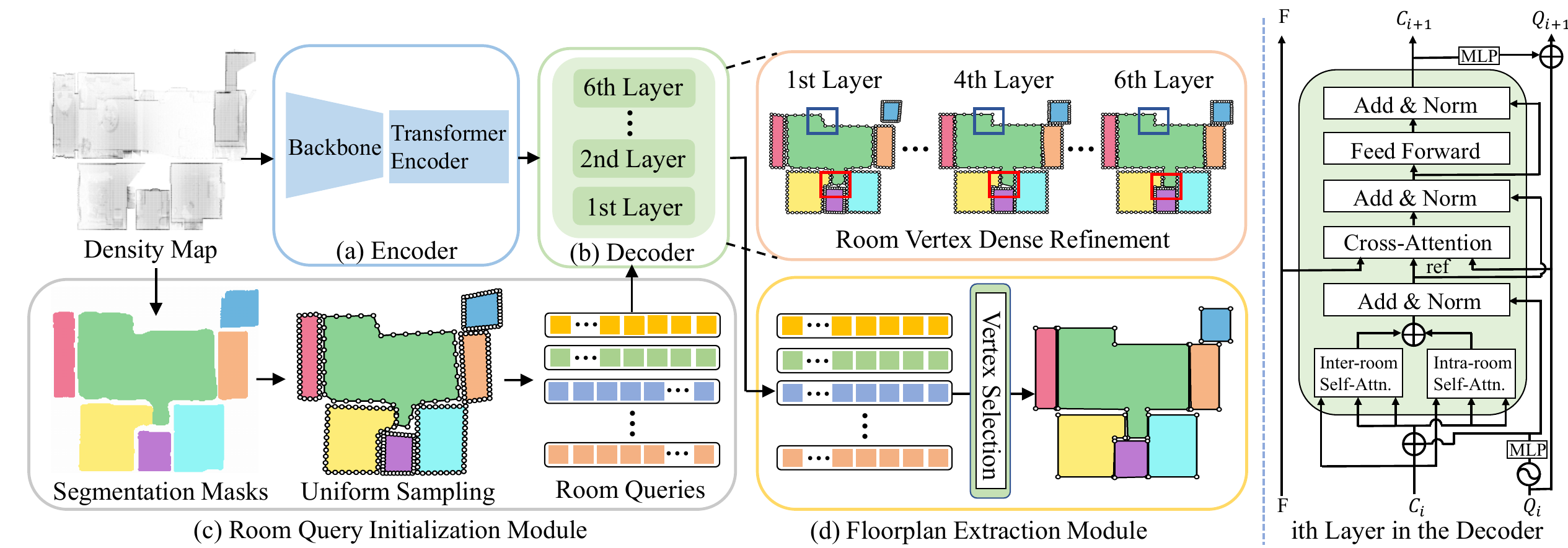}
\caption{\textbf{Overall architecture of PolyRoom.} PolyRoom consists of four main components: (a) Encoder module, (b) Decoder module, (c) Room-aware query initialization module, and (d) Floorplan extraction module. Room queries are initialized with instance segmentation. Subsequently,  they are refined in the Transformer decoder layer by layer with dense supervision (red and blue boxes mark the changes). Finally, the floorplan is extracted based on vertex selection. The detailed structure of the $i$th layer in the Transformer decoder is depicted in the right part, where $F$ denotes the output of the Transformer encoder, $C_i$, $C_{i+1}$ represent content queries from different layers, while $Q_i$, $Q_{i+1}$ denote room queries from different layers.}
\label{pic1}
\end{figure*}
\section{Method}
The proposed PolyRoom views the floorplan reconstruction problem as a progressive refinement process of room queries, which is shown in Figure~\ref{pic1}. In this section, we begin by introducing our floorplan representation, which is followed by the description of the encoder and decoder modules of PolyRoom. Subsequently, we present the room query initialization module and the floorplan extraction module. Finally, we detail the loss functions employed in PolyRoom. 

\subsection{Floorplan Representation}
\begin{figure}[!t] 
\centering
\includegraphics[width=0.95\linewidth]{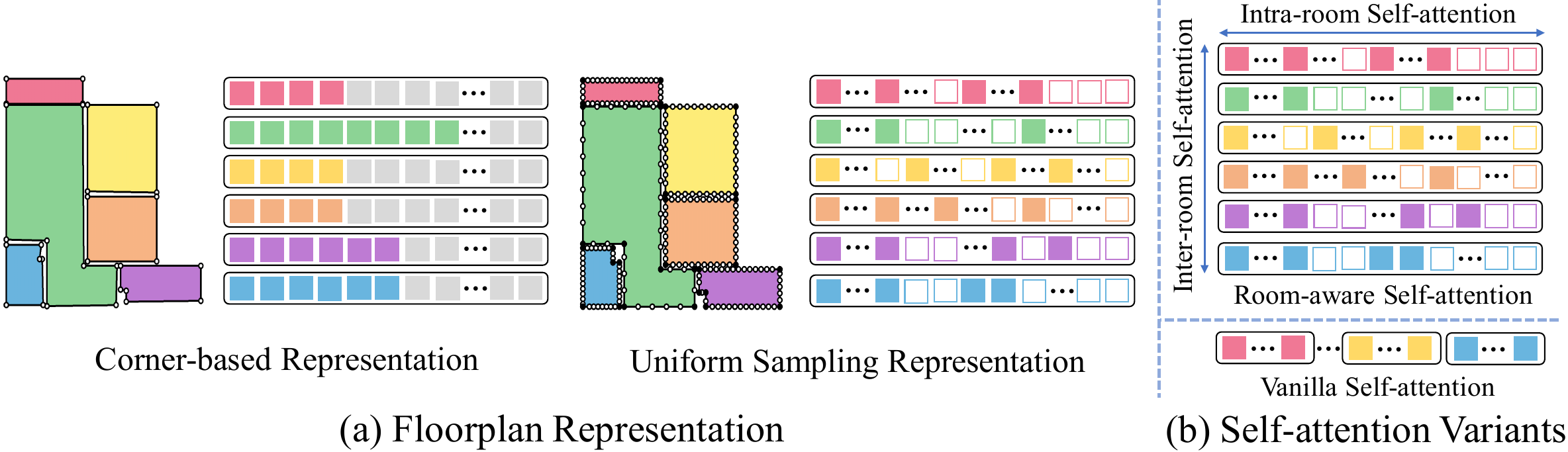}
\caption{(a) Illustration of floorplan representation including the sparse corner-based (left) and our dense uniform sampling representation (right). Valid contour vertices (outlined) and corner vertices (filled) with supervision during training are colored according to the room on the left. (b) Illustration of the self-attention variants including the room-aware self-attention and vanilla self-attention.  Our room-aware self-attention is a combination of intra-room and inter-room self-attention, which works among different vertices in a single room and among different rooms. And the vanilla self-attention performs on the flattened queries.}
\label{pic6}
\end{figure}

In general, since the fundamental structure of a floorplan comprises flat walls connected by angular corners, it is natural and intuitive to represent it with a sequence of 2D polygon vertices~\cite{yue2023connecting,hu2023polybuilding}. As depicted in Figure~\ref{pic6}(a), previous study~\cite{yue2023connecting} has adopted a corner-based sparse floorplan representation, where each polygon possesses a varying number of corners. However,  the sparse supervision may lead to erroneous contour deformations and collapsed topology in room polygons where corners are either missing or biased.   For better supervision of the polygon contours during training,  we opt for a uniform sampling floorplan representation. Here, each polygon is represented by a fixed number of vertices rather than solely relying on corners.  This strategy enables dense supervision along all room walls holistically, thereby bolstering the model's robustness against missing or biased turning corner vertices.

Considering a scene $S$ comprising $M_{gt}$ rooms, the floorplan could be represented as: $S = \{{R}_i\}_{i=1}^{M_{gt}}$, where ${R}_{i}$ denotes the $i$th room.
For a single room, we represent it with  $N$ vertices, which start from the upper-left corner and are arranged in clockwise order: ${R}_{i} =({\mathbf{v}}_{1}^{i},{\mathbf{v}}_{2}^{i},...,{\mathbf{v}}_{j}^{i},...,{\mathbf{v}}_{N}^{i})$, where ${\mathbf{v}_{j}^{i}}$ denotes the $j$th vertex. 
Each vertex is represented by both its position and corner label: ${\mathbf{v}_{j}^{i}}={({\mathbf{p}_{j}^{i}}, l_{j}^{i})}$, where ${\mathbf{p}_{j}^{i}}={(x_{j}^{i}, y_{j}^{i})}$ denotes position coordinate and $l_{j}^{i} \in \{0, 1\}$ indicates whether it is a corner (1) or not (0). To obtain the vertex set, we uniformly sample $N$ vertices along the room contour at equal intervals and set all vertex labels to $0$. Subsequently, for each corner, we select the sampling vertex closest to the corner and replace it with the actual corner, marking it with the label $1$. 
With all corners included, the floorplan is represented by the vertex set.
  \par
In this novel representation, dense supervision is achieved with a fixed number of vertices during training. 
Additionally, the introduction of room-aware query initialization and the utilization of angles are also enabled. 
Because in the corner-based sparse representation, room-aware query initialization becomes unattainable due to the unknown corner number, while angles become highly inaccurate if a certain corner is biased or missing.
\par
\subsection{Encoder}
By projecting an indoor 3D point cloud along the gravity direction, PolyRoom takes the density map generated as input. It then utilizes a CNN backbone~\cite{he2016deep} to extract multi-scale image features.
Then these features from different scales are flattened, concatenated, and added with positional encoding before serving as input to the Transformer  encoder~\cite{vaswani2017attention}. 
Within the encoder, each layer comprises a multi-scale deformable self-attention layer~\cite{zhu2020deformable} and a feed-forward network. 
 Finally, the features output by the encoder interact with queries in the cross-attention layer of the decoder.

\subsection{Decoder}
For the decoder structure, as depicted in Figure~\ref{pic1}, each layer comprises a room-aware self-attention module which combines intra-room self-attention and inter-room self-attention, along with a multi-scale deformable cross-attention module and a feed-forward network. Within each layer, queries initially interact within themselves in the room-aware self-attention module, followed by interaction with the encoder output in the cross-attention module. Ultimately, vertex coordinates are updated through the iterative refinement of room queries layer by layer.

For the queries ($q$), they consist of content queries $C \in \mathbb{R}^{M \times N \times d}$ which are randomly initialized and positional queries $P$ = MLP(PE($Q$)). Here, $M$ denotes the maximum polygon sequence length, $d$ represents the dimension of decoder embedding, MLP refers to a simple multi-layer perception network and PE signifies the positional encoding, which calculates the $d$-dimensional sinusoidal positional embedding of the room queries   $Q\in \mathbb{R}^{M \times N \times 2}$. 

In vanilla self-attention, queries are flattened to interact with each other ($q \in \mathbb{R}^{MN \times d}$), which is constrained by memory consumption with increasing query length. 
Inspired by the self-attention variant in~\cite{liao2023maptrv2}, we introduce the room-aware self-attention, where self-attention is performed along both the intra-room dimension ($q \in \mathbb{R}^{N \times Md}$) and the inter-room dimension ($q \in \mathbb{R}^{M \times Nd}$) (illustrated in Figure~\ref{pic6}(b)). Based on the two-level floorplan representation, the room-aware self-attention focuses on relationships within individual rooms as well as among different rooms. It reduces the computational complexity from $O((MN)^2)$ to $O(M^2+N^2)$, and results in less memory consumption. Moreover,  it also achieves better reconstruction results than vanilla self-attention.

After the room-aware self-attention, the queries interact with the features output by the Transformer encoder in the cross-attention module. 
Following this interaction,  vertex offsets $(\delta x, \delta y)$ are predicted from content queries with an MLP. Upon adding the vertex offsets,  room queries are updated. 
With room queries explicitly represented as polygon vertex coordinates and updated after each decoder layer, room polygons are refined layer by layer. Finally, the polygon vertices are selected to extract the final vectorized floorplan.

\subsection{Room-Aware Query Initialization}
With the aforementioned advantages, the uniform sampling representation also presents a challenge due to longer vertex sequences, making it more difficult for randomly initialized room queries to be accurately learned. This can result in inaccuracies like self-interaction, overlapping, and incorrect sequence orders, as the queries lack a direct correlation to the inputs. Consequently, a room-aware query initialization module becomes imperative. Building upon the new floorplan representation, we propose a method to address the aforementioned challenge by leveraging instance segmentation to generate initial shapes for room queries. 

In detail, the objective of room-aware query initialization is to provide a reasonable estimation of room contours for these queries. To achieve this, room instances are predicted and then represented as room queries.   Specifically,  the pre-trained instance segmentation network first predicts several room masks. 
With the room polygons, starting from the upper-left corner and moving clockwise, we uniformly sample $N$ vertices along the polygon contour at equal intervals. In this way, room queries are explicitly represented as $Q\in \mathbb{R}^{M \times N \times 2}$ with vertex coordinates. If the number of room masks is less than $M$, we use random initial values to pad the room queries. 

\subsection{Floorplan Extraction}

Room queries updated in the Transformer decoder correspond to the predicted polygon contours. For corner prediction, vertex labels are predicted with a feed-forward network from the decoder output. 
In practice, vertices around corners tend to exhibit higher prediction probabilities to be corners, and multiple vertices may be identified as corners based solely on a probability threshold. 
To prevent redundant corners from being selected, we combine prediction probabilities and geometric properties for vertex selection to extract the final floorplan.\par

First, we select vertices predicted to be corners. Then, based on the prior knowledge of indoor buildings that room corner angles are typically not flat or small acute angles,  we also include vertices based on an angle threshold.  Taken together, the conditions for vertex selection can be expressed as:
\begin{align}
\begin{aligned}
pro_{j}&>t_{pro},     \quad
|ang_{j}|&< t_{ang},
\end{aligned}
\end{align}
where $pro_{j}$ is the probability predicted to be corners, $t_{pro}$ is the threshold for probability, $ang_{j}$ indicates the cosine value of the vertex angle and $t_{ang}$ is the threshold for the cosine value.\par

Given the structured and compact nature of floorplan polygons, we utilize a polygonization algorithm~\cite{douglas1973algorithms}  to further select vertices. This ensures that the final vertex sets are minimal, exhibiting robust structural integrity without redundancy, while faithfully preserving the geometry of the original polygon. Subsequently, the final floorplan is extracted.

\subsection{Loss Functions}

\subsubsection{Bipartite Matching.}
To make the network end-to-end trainable, it is supposed to find a bipartite matching between the predicted queries and the ground truth. With the ordered and dense representation, we conduct the bipartite matching totally based on polygon geometry. In detail, we compute distances between the sampling vertices of the predictions and the ground truth, and define the matching cost accordingly as follows:
\begin{align}
\begin{aligned}
C=\sum_{i=1}^{M_{gt}}\sum_{j=1}^{N}d(p_{j}^{i},p_{j}^{\sigma(i)}),
\end{aligned}
\end{align}
where $C$ is the matching cost, $p_{j}^{i}$ and $p_{j}^{\sigma(i)}$ are vertices in the ground truth polygon and the corresponding matching polygon, $d$ is the $L_{1}$ distance between two vertices. All possible arrangements of polygon matching are calculated, and an optimal bipartite matching is assigned with the minimum matching distance. 

\subsubsection{Total Loss.}
After matching, losses can be calculated between the predictions and ground truth. The overall loss function is defined as:
\begin{align}
L =&\sum_{i=1}^{M} (\lambda_{cls}L_{cls}^{i}+\lambda_{coord}L_{coord}^{i} 
+\lambda_{ras}L_{ras}^{i}
+\lambda_{ang}L_{ang}^{i}),
\end{align}
where $L_{cls}$, $L_{coord}$,  $L_{ras}$, and $L_{ang}$  represent the cross-entropy classification loss, $L1$ coordinate loss, rasterization loss and angle loss respectively. $\lambda_{cls}$, $\lambda_{coord}$, $\lambda_{ras}$, and $\lambda_{ang}$ are the corresponding loss weights.  The rasterization loss is also used in RoomFormer~\cite{yue2023connecting}, measuring differences between the reconstructed and GT shapes through differentiable rendering. The angle loss calculates angle differences  to better supervise  the densely sampled vertex positions:
\begin{align}
L_{ang}^{i}=\frac{1}{N}\sum_{j=1}^{N}|ang_{j}^{i} - ang_{j}^{\sigma(i)}|, i \leq M_{gt},
\end{align}
where $ang_{j}^{i}$ and $ang_{j}^{\sigma(i)}$ represent the cosine value of the vertex angle in the ground truth and matching polygon respectively.\par

\section{Experiments}
\subsection{Datasets and Evaluation Metrics}
\subsubsection{Datasets.}
We evaluate our proposed method in two popular indoor environment datasets, Structured3D~\cite{zheng2020structured3d} and SceneCAD~\cite{avetisyan2020scenecad}. Structured3D is a synthetic dataset with 3D structure annotations. There are 3500 houses in the dataset (3000/250/250 for training/validation/test) containing rich floorplan types. SceneCAD provides CAD-based representations of scanned 3D environments~\cite{dai2017scannet} from commodity RGB-D sensors, and floorplan annotations are extracted from them.  The data of both the two datasets are projected along the vertical axis into 256×256 top-view point-density images as inputs.\par 
\subsubsection{Evaluation Metrics.}
For evaluation metrics, we adopt the same criteria as previous works~\cite{yue2023connecting,chen2022heat}, including three-level metrics: Room, Corner, and Angle. Each level metric comprises Precision, Recall, and F1 score. Additionally, for the Room level metric, Intersection over Union (IoU) is reported.
\subsection{Implementation Details}
We utilize ResNet-50~\cite{he2016deep} as our backbone and pre-train Mask2Former~\cite{cheng2022masked} on each dataset for 200 epochs as our instance segmentation network. We set the maximum number of rooms as 20 and the number of sampling vertices as 40. For the loss weights, we assign the values of $\lambda_{cls}$, $\lambda_{coord}$, $\lambda_{angle}$, and $\lambda_{ras}$ as 2, 5, 1, and 1.  For vertex selection, $t_{pro}$ is set to 0.01, $t_{ang}$ is set to $\sqrt{3}/{2}$, and the DP threshold is set to 4 for the Structured3D dataset and 10 for the  SceneCAD dataset. During training, the model is trained on Structured3D for 500 epochs with a learning rate of 2e-4 and on SceneCAD for 400 epochs with a learning rate of 5e-5. All experiments are optimized by Adam~\cite{kingma2014adam} and implemented in pytorch with a batch size of 40 on an NVIDIA RTX 3090 GPU.

\begin{table*}[!t]
\caption{\textbf{Quantitative evaluation results on Structured3D~\cite{zheng2020structured3d} dataset.} Results of previous works are taken from~\cite{yue2023connecting} and~\cite{su2023slibo}. The colors \textcolor{orange}{\textbf{orange}} and \textcolor{cyan}{\textbf{cyan}} mark the top-two results, where~$^{*}$ indicates the use of our vertex selection approach.}
\setlength{\tabcolsep}{10pt} 
\centering
\resizebox{\textwidth}{!}{
\begin{tabular}{@{}lllllllllll@{}}
\toprule
{} & {}& \multicolumn{3}{c}{Room}                   & \multicolumn{3}{c}{Corner}                 & \multicolumn{3}{c}{Angle}                   \\ \cmidrule(l){3-5} \cmidrule(l){6-8} \cmidrule(l){9-11}
Method  & Presented at         & Prec.        & Rec.         & F1           & Prec.        & Rec.         & F1           & Prec.        & Rec.         & F1           \\ \midrule
DP      &-        & 96.0           & 95.0           & 95.5           & 82.8           & 78.6           & 80.7           & 57.3           & 54.3           & 55.8           \\
Floor-SP~\cite{chen2019floor}   & ICCV19     & 89.          & 88.          & 88.          & 81.           & 73.           & 76.           & 80.           & 72.           & 75.           \\
MonteFloor~\cite{stekovic2021montefloor}  &ICCV21    & 95.6         & 94.4         & 95.0         & 88.5         & 77.2         & 82.5         & 86.3         & 75.4         & 80.5         \\ \midrule
HAWP~\cite{xue2020holistically}       &CVPR20     & 77.7         & 87.6         & 82.3         & 65.8         & 77.0         & 70.9         & 59.9         & 69.7         & 64.4         \\
LETR~\cite{xu2021line} &CVPR21 & 94.5         & 90.0         & 92.2         & 79.7         & 78.2         & 78.9         & 72.5         & 71.3         & 71.9         \\
HEAT~\cite{chen2022heat}  &CVPR22          & 96.9         & 94.0         & 95.4         & 81.7         & 83.2         & 82.5         & 77.6         & 79.0         & 78.3         \\
RoomFormer~\cite{yue2023connecting}   &CVPR23   & 97.9         & 96.7         & 97.3         & 89.1         & \textcolor{cyan}{\textbf{85.3}}         & 87.2        & 83.0         & 79.5         & 81.2         \\ 
RoomFormer$^{*}$ ~\cite{yue2023connecting}   &CVPR23   & 97.8         & 96.2         & 97.0         & \textcolor{cyan}{\textbf{91.6}}         & 83.8         & \textcolor{cyan}{\textbf{87.5}}         & 86.1         & 78.9        & 82.3         \\ 
SLIBO-Net~\cite{su2023slibo}   &NeurIPS23   & \textcolor{orange}{\textbf{99.1}}         & \textcolor{orange}{\textbf{97.8}}         & \textcolor{orange}{\textbf{98.4}}         & 88.9         & 82.1         & 85.4         & \textcolor{cyan}{\textbf{87.8}}         & \textcolor{cyan}{\textbf{81.2}}         & \textcolor{cyan}{\textbf{84.4}}         \\ \midrule
PolyRoom (Ours) & - & \textcolor{cyan}{\textbf{98.9}} & \textcolor{cyan}{\textbf{97.7}} & \textcolor{cyan}{\textbf{98.3}} & \textcolor{orange}{\textbf{94.6}} & \textcolor{orange}{\textbf{86.1}} & \textcolor{orange}{\textbf{90.2}} & \textcolor{orange}{\textbf{89.3}} & \textcolor{orange}{\textbf{81.4}} & \textcolor{orange}{\textbf{85.2}} \\ \bottomrule

\end{tabular}}
\label{Table:1}
\end{table*}

\begin{table*}[!t]
\caption{\textbf{Quantitative evaluation results on SceneCAD~\cite{avetisyan2020scenecad} dataset.}  The colors \textcolor{orange}{\textbf{orange}} and \textcolor{cyan}{\textbf{cyan}} mark the top-two results, where~$^{*}$ indicates the use of our vertex selection approach.
}
\setlength{\tabcolsep}{10pt}
\centering
\resizebox{0.7\textwidth}{!}{
\begin{tabular}{@{}llllllll@{}}
\toprule
               & \multicolumn{1}{c}{Room}         & \multicolumn{3}{c}{Corner}                 & \multicolumn{3}{c}{Angle}                  \\ \cmidrule(l){2-2} \cmidrule(l){3-5} \cmidrule(l){6-8}
Method         & IoU          & Prec.        & Rec.         & F1           & Prec.        & Rec.         & F1           \\ \midrule
DP             & 90.1           & 62.4           & 84.3           & 71.7           & 24.2           & 32.3           & 27.7          \\
Floor-SP       & 91.6         & 89.4         & \textcolor{cyan}{\textbf{85.8}}         & 87.6         & 74.3         & 71.9         & 73.1         \\
HEAT           & 84.9         & 87.8         & 79.1         & 83.2         & 73.2         & 67.8         & 70.4         \\
RoomFormer    & \textcolor{cyan}{\textbf{91.7}}         & 92.5        & 85.3         & 88.8         & 78.0         & \textcolor{cyan}{\textbf{73.7}}         & 75.8 \\
RoomFormer$^{*}$    & 91.3        & \textcolor{cyan}{\textbf{94.4}}         & 84.7         & \textcolor{cyan}{\textbf{89.2}}         & \textcolor{cyan}{\textbf{79.6}}         & 73.4         & \textcolor{cyan}{\textbf{76.4}} 

\\ \midrule
PolyRoom (Ours) & \textcolor{orange}{\textbf{92.8}} & \textcolor{orange}{\textbf{96.8}} & \textcolor{orange}{\textbf{86.1}} & \textcolor{orange}{\textbf{91.2}} & \textcolor{orange}{\textbf{81.7}} & \textcolor{orange}{\textbf{74.5}} & \textcolor{orange}{\textbf{78.0}} \\ \bottomrule
\end{tabular}}
\label{Table:2}
\end{table*}
\subsection{Comparison with State of the Arts}
\subsubsection{Quantitative Evaluations.}
We compare our method with eight other approaches, including DP~\cite{douglas1973algorithms}, Floor-SP~\cite{chen2019floor}, MonteFloor~\cite{stekovic2021montefloor}, HAWP~\cite{xue2020holistically}, LETR~\cite{xu2021line}, HEAT~\cite{chen2022heat}, RoomFormer~\cite{yue2023connecting}, and SLIBO-Net~\cite{su2023slibo}. 
DP simplifies room segments with DP~\cite{douglas1973algorithms} polygonization algorithm. 
Floor-SP and MonteFloor are optimization algorithms with room segmentation. 
HAWP and LETR  are general methods for wireframe parsing and line segment detection. 
HEAT, RoomFormer, and SLIBO-Net are Transformer-based methods for floorplan reconstruction. \par
On Structured3D (Table~\ref{Table:1}), PolyRoom is compared with all the comparative methods, and on SceneCAD (Table~\ref{Table:2}), it is compared with four representative ones, including DP, Floor-SP, HEAT, and RoomFormer. On both datasets, PolyRoom outperforms all other comparative methods. Notably, on Structured3D, PolyRoom enhances the F1 score of corner and angle by 4.8\% and 0.8\% respectively compared to the SLIBO-Net which relies on the Manhattan assumption, while maintaining a similar result in the F1 score of the room. In comparison to RoomFormer, PolyRoom surpasses it across all three metrics significantly. On SceneCAD, PolyRoom achieves notable improvements, with increases of 1.1\%, 2.0\%, and 1.6\% for Room IoU, Corner F1 score, and Angle F1 score respectively. \par

\begin{figure}[!t] 
\begin{minipage}{\textwidth}
\centering
\includegraphics[width=\linewidth]{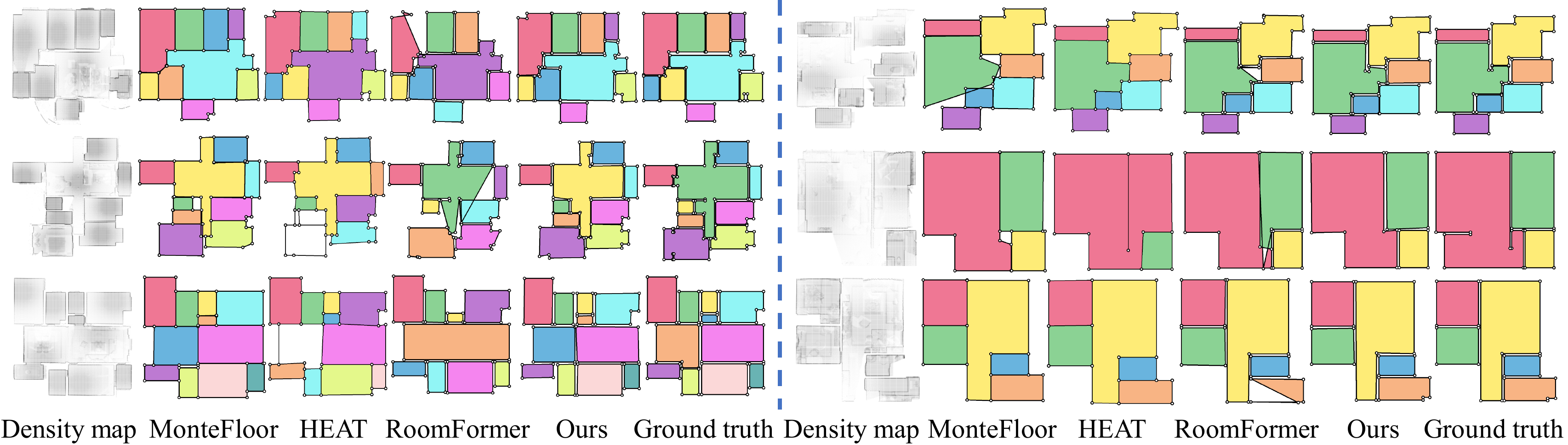}
\captionsetup{width=\textwidth} 
\caption{\textbf{Qualitative evaluations on Structured3D~\cite{zheng2020structured3d}}. PolyRoom performs with fewer missing rooms, more correct room sequences, and improved room details.}
\label{pic4}
\end{minipage}
\end{figure}
\begin{figure}[!t] 
\centering
\begin{minipage}{\textwidth}
\centering
\includegraphics[width=\linewidth]{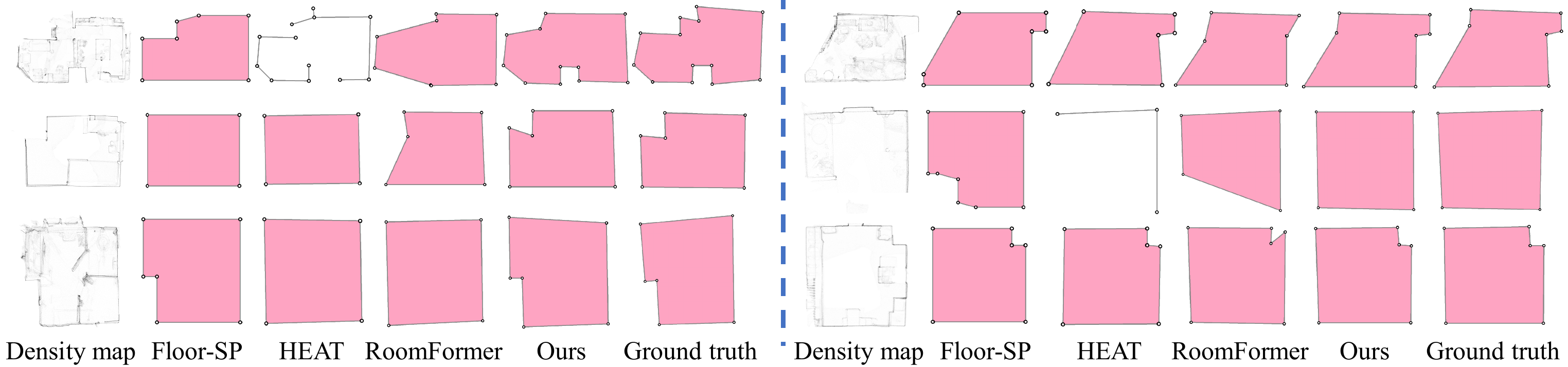}
\caption{\textbf{Qualitative evaluations on SceneCAD~\cite{avetisyan2020scenecad}}. }
\label{pic5}
\end{minipage}
\end{figure}

\subsubsection{Qualitative Evaluations.}
Some results on Structured3D and SceneCAD are shown in Figure~\ref{pic4} and  Figure~\ref{pic5} for qualitative comparison, it can be observed that MonteFloor and Floor-SP exhibit incorrect shapes of rooms, HEAT encounters issues with missing corners and edges, resulting in unclosed room polygons. RoomFormer suffers from missing rooms and inaccurate polygon sequences, such as self-intersection and overlap of polygons. In contrast, PolyRoom delivers more complete reconstruction results with fewer missing rooms, more accurate polygon sequences, and improved reconstruction details. 

\subsubsection{Cross-data Generalization.}
We evaluate the cross-data generalization ability of PolyRoom across different datasets.  Our model is trained on the Structured3D training set and then tested on the SceneCAD validation set. We compare the results with two end-to-end competing floorplan reconstruction methods: HEAT and RoomFormer. The findings presented in Table~\ref{Table3} demonstrate that PolyRoom exhibits superior generalization compared to other methods, owing to the proposed room-aware query initialization and dense supervision mechanisms.
\begin{table}[!t]
\caption{\textbf{Cross-data generalization}. The best results are in bold font, the same as in the following tables.}
\setlength{\tabcolsep}{10pt}
\centering
\resizebox{0.7\textwidth}{!}{
\begin{tabular}{@{}llllllll@{}}
\toprule
               & \multicolumn{1}{c}{Room}         & \multicolumn{3}{c}{Corner}                 & \multicolumn{3}{c}{Angle}                  \\ \cmidrule(l){2-2} \cmidrule(l){3-5} \cmidrule(l){6-8}
Method         & IoU          & Prec.        & Rec.         & F1           & Prec.        & Rec.         & F1           \\ \midrule
HEAT           & 52.5         & 50.9         & 51.1         & 51.0         & 42.2         & 42.0         & 41.6         \\
RoomFormer  & 74.0         & 56.2         & 65.0         & 60.3         & 44.2         & 48.4 & 46.2         \\ \midrule
PolyRoom (Ours) & \textbf{85.2} & \textbf{77.8} & \textbf{79.9} & \textbf{78.9} & \textbf{59.4} & \textbf{61.7} & \textbf{60.6} \\ \bottomrule
\end{tabular}}
\label{Table3}
\end{table}

\begin{table*}[!t]
\centering
\caption{\textbf{Ablation study results of different components of PolyRoom on Structure3D~\cite{zheng2020structured3d}.}}
\setlength{\tabcolsep}{10pt} 
\resizebox{0.82\textwidth}{!}{
\begin{tabular}{@{}lllllll@{}}
\toprule
& \multicolumn{2}{c}{Room}                   & \multicolumn{2}{c}{Corner}                 & \multicolumn{2}{c}{Angle}              \\ \cmidrule(l){2-3} \cmidrule(l){4-5} \cmidrule(l){6-7}
Settings         & Prec.        & Rec.                 & Prec.        & Rec.                  & Prec.        & Rec.               \\ \midrule
Baseline & \textbf{98.9} & \textbf{97.7}& \textbf{94.6} & 86.1 & \textbf{89.3} & \textbf{81.4}           \\
w/o Uniform sampling representation & 86.7         & 70.4         & 78.2        & 61.2         & 71.8         & 56.2                    \\
w/o Room-aware query initialization & 96.9         & 96.4        & 89.3        & 84.6        & 82.7         & 78.4                 \\
w/o Room-aware self-attention & 98.8         & 97.4        & 94.1         & 85.8         & 89.0         & 81.2                   \\
w/o Vertex selection               
& 98.7         & 97.6         & 81.4         & \textbf{87.9}         & 72.6                  & 77.9 \\
\bottomrule
\end{tabular}}
\label{Table: 4}
\end{table*}

\subsection{Ablation Studies}
\subsubsection{Different Components.}
Table~\ref{Table: 4} illustrates the impacts of different components on the final results. 
Without uniform sampling representation, room-aware query initialization introduces room initial shapes with varying vertex numbers compared to the ground truth, posing a challenge in predicting accurate corner counts. 
Conversely, without room-aware query initialization, employing a single uniform sampling representation results in a higher number of vertices, hindering the accurate regression of vertex coordinates and leading to inferior performance.
 Additionally, room-aware self-attention exhibits superior performance compared to vanilla self-attention.
Regarding vertex selection, it removes redundant vertices, yielding more accurate and structured results. 
\begin{table}[!t]
\centering
\caption{\textbf{Analysis on layer number and vertex number of PolyRoom on Structure3D~\cite{zheng2020structured3d}.} }
\setlength{\tabcolsep}{10pt} 
\resizebox{0.65\textwidth}{!}{
\begin{tabular}{@{}llllllll@{}}
\toprule
\multicolumn{2}{c}{Settings}    & \multicolumn{2}{c}{Room}   & \multicolumn{2}{c}{Corner}   & \multicolumn{2}{c}{Angle}                   \\ \cmidrule(r){1-2} \cmidrule(l){3-4} \cmidrule(l){5-6} \cmidrule(l){7-8}
Layer & Vertex & Prec.        & Rec.              & Prec.        & Rec.                & Prec.        & Rec.              \\ \midrule
6 & 40          &      \textbf{98.9} & 97.7& 94.6 & 86.1 & \textbf{89.3} & \textbf{81.4}   \\ \midrule
4 & 40                        &   98.2       &    96.9 & 93.5&85.0 & 87.9 &80.1              \\
8 & 40       &98.7     &97.6      &   \textbf{94.7}      &    86.0     &   89.2&81.3      \\

\toprule
6 & 30    &  98.2 &97.2 &93.9 & 85.9 & 87.8 & 80.4            \\
6 & 50   &   98.8     &   \textbf{97.8}       &     94.4    &  \textbf{86.3} & 87.6&80.2           \\ 
 \bottomrule
\end{tabular}}
\label{Table: 5}
\end{table}
\subsubsection{Vertex and Layer Number.}
 In this experiment, we investigate the effects of vertex number and layer number on the final results.  As depicted in Table~\ref{Table: 5}, within a certain range, as the layer number and vertex number increase, the model performance increases, while a high layer number yields similar results, and a high vertex number results in a decrease in performance. To strike a balance between accuracy and training time,  we select a layer number of 6 and a vertex number of 40 as our parameters.

\subsubsection{Robustness to Initialization.}
 To further investigate the impact of initialization on the final results, we compare different instance segmentation networks, including the simple Mask-RCNN~\cite{he2017mask} and state-of-the-art  Mask2Former~\cite{cheng2022masked} with different backbones. 
Table~\ref{Table: 6} presents the reconstruction results based on different segmentation qualities (measured by mean Average Precision (mAP)) and model sizes (measured by parameters (Para.)). 
Enhanced instance segmentation leads to improved room-aware query initialization, resulting in superior reconstruction results. Additionally, the final results also demonstrate robustness to segmentation results of different qualities.
\begin{table}[!t]
\caption{\textbf{Robustness to initialization on Structure3D~\cite{zheng2020structured3d}.}  }
\centering
\setlength{\tabcolsep}{10pt}
\resizebox{0.9\textwidth}{!}{
\begin{tabular}{@{}lllllllllllll@{}}
\toprule
 & & & & \multicolumn{2}{c}{Room}                   & \multicolumn{2}{c}{Corner}                 & \multicolumn{2}{c}{Angle}                   \\  \cmidrule(l){5-6} \cmidrule(l){7-8} \cmidrule(l){9-10}
 Network& Backbone&Para.(M) & mAP        & Prec.        & Rec.               & Prec.        & Rec.                & Prec.        & Rec.              \\ \midrule
Mask-RCNN & Res50    &44.1         & 39.3           & 98.6          & 97.3           & 92.6          & 86.0    &87.0       & 81.1       \\
Mask2Former &  Res50  &44.0   &41.6        & 98.6           & 97.4           & 93.0           & 85.9          & 87.5           & 81.1       \\
 Mask2Former & Res101 &63.0 & 44.3                  & 98.7        & 97.4     &94.0    & 86.0         & 88.3        & 81.2        \\ 
 Mask2Former & Swin-S  & 68.7& 45.3                  & 98.7        & 97.4         & 94.2         & 86.0        & 88.8 &81.2         \\ 
 Mask2Former & Swin-L     &215.5 &48.0          & \textbf{98.9} & \textbf{97.7}& \textbf{94.6} & \textbf{86.1} & \textbf{89.3} & \textbf{81.4}       \\
 \bottomrule
\end{tabular}}
\label{Table: 6}
\end{table}
\begin{table}[!t]
\caption{\textbf{Ablation on the self-attention variants on Structure3D~\cite{zheng2020structured3d}.} The GPU memory is tested with a batch size of 20, as the model with vanilla self-attention runs out of memory with a batch size of 40. }
\centering
\setlength{\tabcolsep}{10pt}
\resizebox{0.73\textwidth}{!}{
\begin{tabular}{@{}llllllll@{}}
\toprule
&  & \multicolumn{2}{c}{Room}                   & \multicolumn{2}{c}{Corner}                 & \multicolumn{2}{c}{Angle}                   \\  \cmidrule(l){3-4} \cmidrule(l){5-6} \cmidrule(l){7-8}
Self-attention & Memory        & Prec.        & Rec.              & Prec.        & Rec.                 & Prec.        & Rec.               \\ \midrule
Vanilla &18707M          & 98.8         & 97.4        & 94.1         & 85.8         & 89.0         & 81.2            \\
Single intra-room & 11071M              &   98.2         &    96.9        &      93.3      &   84.9        &     88.2      &   80.4     \\
Single inter-room & 10947M   &  98.6                 &   97.3      &   94.3      &     85.7    &   88.6     &       80.8  \\ 
Room-aware & 11957M                & \textbf{98.9} & \textbf{97.7}& \textbf{94.6} & \textbf{86.1} & \textbf{89.3} & \textbf{81.4}        \\
 \bottomrule
\end{tabular}}
\label{Table: 7}
\end{table}
\begin{table}[!t]
\caption{\textbf{Loss function and vertex selection analysis on Structure3D~\cite{zheng2020structured3d}.} }
\centering
\setlength{\tabcolsep}{10pt} 
\resizebox{0.67\textwidth}{!}{
\begin{tabular}{@{}lllllll@{}}
\toprule
& \multicolumn{2}{c}{Room}                   & \multicolumn{2}{c}{Corner}                 & \multicolumn{2}{c}{Angle}                   \\ \cmidrule(l){2-3} \cmidrule(l){4-5} \cmidrule(l){6-7}
 Settings          & Prec.        & Rec.            & Prec.        & Rec.                  & Prec.        & Rec.               \\ \midrule
Ours & 98.9 & 97.7& 94.6 & 86.1 & 89.3 & 81.4         \\ \midrule
w/o Angle loss            & 98.1        & 96.5           & 92.6           & 85.2           & 85.0          & 78.3           \\  \midrule
w/o Angle threshold            & 98.7        & 97.6           & 93.8           & 86.3           & 88.2          & 81.1           \\  
w/o Polygonization           & 98.8        & 97.7           & 92.6          & 87.0           & 86.7          & 81.5           \\  
\bottomrule
\end{tabular}}
\label{Table: 8}
\end{table}
\subsubsection{Self-attention Variants.}
In Table~\ref{Table: 7}, we conduct a detailed ablation experiment to validate the effectiveness of room-aware self-attention. 
It demonstrates much less training memory compared to vanilla self-attention while also achieving superior performance.
Furthermore, compared to using intra-room and inter-room self-attention separately, room-aware self-attention proves to be more suitable for our floorplan representation and achieves better performance.

\subsubsection{Loss Functions and Vertex Selection.}
In this experiment, we evaluate the necessity of angle loss, angle threshold, and polygonization in vertex selection.
Angle loss serves as an auxiliary loss focusing on angle perspective to better supervise vertex positions.
Table~\ref{Table: 8} illustrates that the introduction of angle loss improves reconstruction results, particularly benefiting angle prediction.  Concerning vertex selection, both angle threshold and polygonization demonstrate a positive impact in reducing redundant vertices.
\begin{figure}[!t] 
\centering
\includegraphics[width=\linewidth]{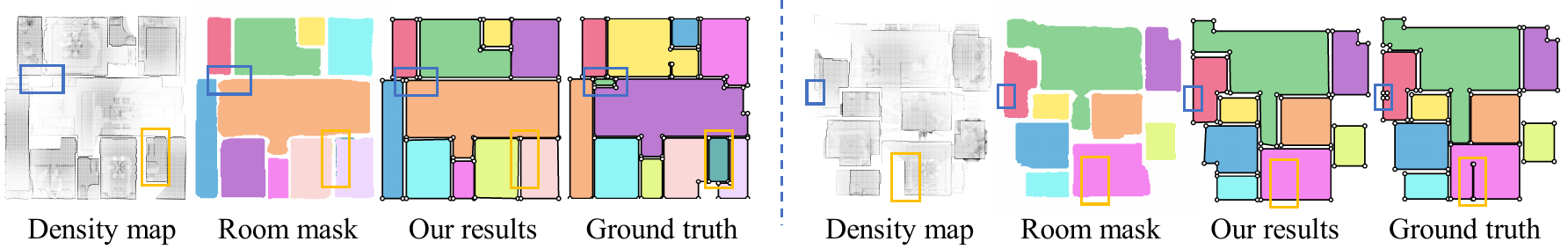}
\caption{\textbf{Failure cases.} (Left) Our model misses rooms because of poor initialization. (Right) Our model misses details when dealing with complex structures. (Blue and orange boxes mark areas where reconstruction fails.)}
\label{pic3}
\end{figure}
\subsection{Analysis of Failure Cases}
Despite exhibiting improved robustness with segmentation networks of varying qualities, our model still encounters difficulty in accurately predicting the complete floorplan when certain room instances are missing from the segmentations  (left in Figure~\ref{pic3}). Moreover, when dealing with complex indoor structures, PolyRoom could overlook details such as thin walls, which can be attributed to limitations in segmentation results and raw data quality (right in Figure~\ref{pic3}).

\section{Conclusion}
In this paper, we present PolyRoom, a room-aware Transformer for floorplan reconstruction. Utilizing uniform sampling representation, room-aware query initialization, and room-aware self-attention, PolyRoom offers significant advancements in the field.  With room-aware query initialization, we establish a direct link between room semantics and room query positions, enhancing the reconstruction accuracy.  Subsequently, room queries are progressively refined in the decoder layer by layer, benefiting from dense supervision throughout. The final floorplan is derived through vertex selection, incorporating both probabilistic considerations and geometric constraints. Moreover, the adoption of room-aware self-attention not only reduces memory consumption but also improves reconstruction quality.  Both qualitative and quantitative evaluations demonstrate the superiority of PolyRoom over other existing methods while with better generalization. Our comprehensive ablation study further validates the efficacy of our proposed design choices.  We hope our work will inspire further research on reconstruction methods with polygon sequences.
\section*{Acknowledgements}
We thank the authors of RoomFormer, HEAT and MonteFloor for providing experimental results, especially Yuanwen Yue for his kind help with the experiments. This work was supported by the National Natural Science Foundation of China (No. U22B2055, 62273345, and 62373349), the Beijing Natural Science Foundation (No. L223003), and the Key R\&D Project in Henan Province (No. 231111210300).

%
%
\bibliographystyle{splncs04}
\bibliography{main}
\end{document}